\documentclass[letterpaper, 10 pt, conference]{ieeeconf}  

\IEEEoverridecommandlockouts                              

\let\labelindent\relax
\usepackage{amsmath} 
\usepackage{amssymb}  
\usepackage{graphicx}
\usepackage[export]{adjustbox}
\usepackage{array}
\usepackage{multirow}
\usepackage{tabularx}
\usepackage{subcaption}
\usepackage[utf8]{inputenc}
\usepackage[export]{adjustbox}
\usepackage{wrapfig}
\usepackage{threeparttable}
\usepackage[english]{babel}
\usepackage{color, colortbl}
\usepackage{makecell}

\usepackage{xpatch}
\usepackage{array}
\usepackage{multirow}
\usepackage{graphicx}
\usepackage{colortbl}
\usepackage{pifont}
\usepackage{enumitem}
\newcommand{\cmark}{\ding{51}}%

\usepackage{overpic}
\usepackage{booktabs}
\usepackage[table]{xcolor}
\definecolor{mycitecolor}{RGB}{0, 0.4, 0.7}
\definecolor{mypink1}{RGB}{139, 0, 0}
\usepackage[breaklinks=true,bookmarks=true,citecolor=mycitecolor,colorlinks,pagebackref=true]{hyperref}
\usepackage[super]{nth}

\title{\LARGE \bf
Sparse Dense Fusion for 3D Object Detection
}

\author{Yulu Gao, Chonghao Sima, Shaoshuai Shi, Shangzhe Di, Si Liu and Hongyang Li
\thanks{Yulu Gao, Shangzhe Di, and Si Liu are with Institute of Artificial Intelligence, Beihang University, Beijing, China.  
Chonghao Sima and Hongyang Li are with OpenDriveLab, Shanghai AI Laboratory, Shanghai, China.
Shaoshuai Shi is with the Department of Electronic
Engineering at The Chinese University of Hong Kong, Hong Kong, China.
 }%
}

\begin{document}

\maketitle
\thispagestyle{empty}
\pagestyle{empty}

\begin{abstract}

With the prevalence of multimodal learning, camera-LiDAR fusion has gained popularity in 3D object detection. 
   Although multiple fusion approaches have been proposed, they can be  
   classified into either sparse-only or dense-only fashion based on the feature representation in the fusion module.
   In this paper, we {analyze them in a common taxonomy} and thereafter observe two challenges: 1) sparse-only solutions preserve 3D geometric prior and yet lose rich semantic information from the camera, and 2) dense-only alternatives retain the semantic continuity but miss the accurate geometric information from LiDAR.
   By analyzing these two formulations, we conclude that the information loss is inevitable due to their design scheme.
   To compensate for the information loss in either manner, we propose Sparse Dense Fusion (SDF), a complementary framework that incorporates both sparse-fusion and dense-fusion modules via the Transformer architecture. Such a simple yet effective sparse-dense fusion structure enriches semantic texture and exploits spatial structure information simultaneously. Through our SDF strategy, we assemble two popular methods with moderate performance and outperform baseline by \textbf{4.3\%} in \textit{mAP} and \textbf{2.5\%} in NDS, ranking first on the nuScenes benchmark. Extensive ablations demonstrate the effectiveness of our method and empirically align our analysis.

\end{abstract}

\section{INTRODUCTION}

The human experience of the world is multimodal - through vision, auditory, thermoception, and other sensory systems.
Modality denotes how something happens or is experienced and multimodal learning is characterized as a research problem when a machine learning system includes multiple such modalities~\cite{baltruvsaitis2018multimodal}.
This problem comes of vital importance in the scenario of autonomous driving as sensor configurations get more complex on vehicles, providing a suitable playground~\cite{caesar2020nuscenes}. 
Within the realm of 3D object detection, point cloud from LiDAR and RGB images from camera are two general perception sources with complementarity and redundancy to each other.
While point cloud can provide accurate 3D spatial structure around the ego vehicle, RGB images perceive the scene with rich semantic information.

\begin{figure}
  \centering
      \includegraphics[width=1.0\columnwidth]{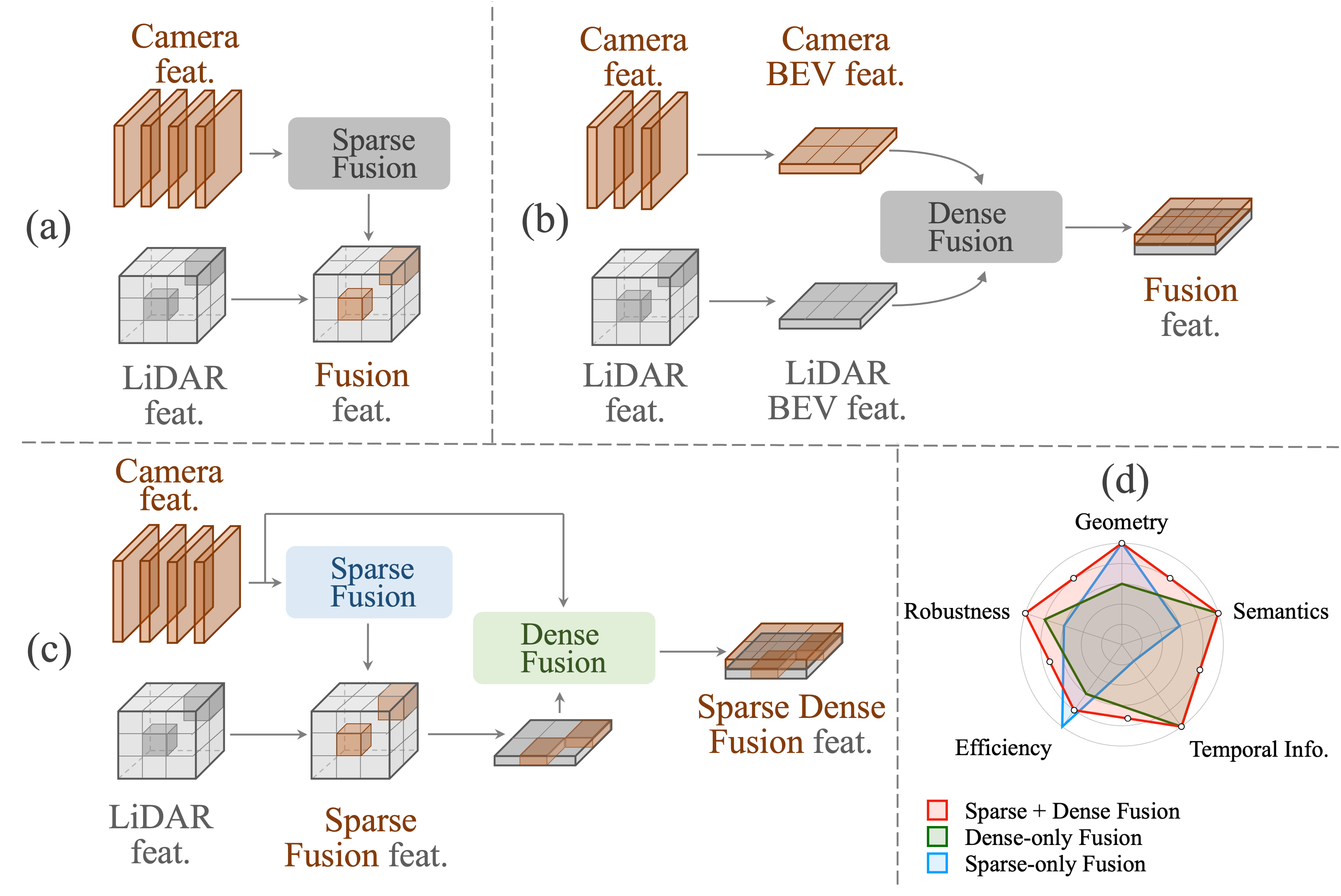}
  \caption{\textbf{Design scheme} of (a) sparse-only fusion, preserving 3D geometric prior and yet losing rich semantic information, (b) dense-only fusion, retaining the semantic continuity but missing the accurate geometric information, and (c) our sparse-dense fusion, preserving 3D geometric prior and semantic information simultaneously. (d) illustrates the strengths and weaknesses of each fashion and the corresponding design scheme. The difference in geometry, semantics, and temporal information is demonstrated via mAP and NDS in Tab.~\ref{tb:components}, while the difference in robustness  explained in Tab.~\ref{tb:robustness} and efficiency is explained in Tab.~\ref{tb:efficiency}.}
  \label{fig:motivation}

  \vspace{-0.25cm}
\end{figure}

Given such a complementary setting of sensors, multimodal learning naturally raises the audience's interest in information interaction between those sensors.
However, the design is non-trivial due to the considerable difference between the sparsity in LiDAR data and the density in camera data.
Distinguished by the representation of feature in fusion modules, we categorize existing methods into two fashion: \textbf{\textit{sparse-only}} fusion
and \textbf{\textit{dense-only}} fusion.
To alleviate the sparsity in LiDAR data, \textbf{\textit{sparse-only}} methods usually transform image semantic label~\cite{vora2020pointpainting,wang2021pointaugmenting} or feature~\cite{chen2022autoalign,chen2022autoalignv2,liu2021epnet++} to point cloud representation via LiDAR-camera sampling~\cite{liang2019multi,Yin2021MVP} or predefined depth bins~\cite{wu2022sparse,zhu2022vpfnet}.
To address the density in camera data, intuitive approaches in \textbf{\textit{dense-only}} fashion are fusing feature from both modalities in bird's-eye-view (BEV) representation~\cite{liu2022bevfusion,liang2022bevfusion,yang2022deepinteraction,yoo20203d,li2022deepfusion,bai2022transfusion}.
Besides BEV, voxel representation also draws some interest in dense-only fusion.
The camera-voxel transformation utilizes either camera parameters~\cite{li2022uvtr} or estimated depth per pixel~\cite{li2022homogeneous} to establish the projection mapping.
As we observe this vast difference between the two fashions, some natural but non-trivial questions are raised - 
\emph{what is the core issue left unexplored in sensor fusion? what are the key factors to improve the upper bound of performance?
}

We conduct a caveat analysis of the prevailing sensor fusion methods, and it reveals that either fashion has inherent advantages and disadvantages due to its design scheme.
Fig.~\ref{fig:motivation} (d) (a) (b) illustrates the strengths and weaknesses of each fashion and corresponding design scheme in a nutshell.

In the \textbf{\textit{sparse-only}} fashion, point cloud is projected to image plane and gather semantic label or image feature around the projection position, thus fusing image information to point cloud representation.
Though this can preserve 3D geometric information in the point cloud feature, the rich semantic information provided by the image feature is broken.
This is because a 3D object, described by several points in point cloud data, corresponds to hundreds of pixels in image data.
This sparsity-density inconsistency leads to a problem: when model appends image feature to voxel representation, around 1 to 2 magnitude of image feature is deprecated via LiDAR-camera projection.

In \textbf{\textit{dense-only}} fashion, image feature and point cloud feature are transformed to the BEV space, respectively, then fused grid by grid there.
While transforming image feature to BEV space remains semantic information, squeezing point cloud feature to BEV space genuinely loses the geometric prior in 3D space.
Also, projecting image feature to BEV space is ill-posed since camera does not capture any 3D geometry.
This results in the grid inconsistency of the same object between image-BEV and point-cloud-BEV feature, increasing the difficulty of fusion.
Therefore, information loss is inevitable in either manner, and we are motivated to design a fusion mechanism that sparse-only and dense-only fusion can jointly contribute to the network.

Given that the two {fashions operate feature map in different representations (voxel, perspective and BEV)}, Transformer architecture~\cite{vaswani2017attention} meets our demands in aggregating valuable information from different modalities conceptually.
Therefore, we propose \textbf{{Sparse-Dense~Fusion}} (\textbf{{SDF}}), a complementary structure connecting sparse-fusion and dense-fusion modules to enjoy the best of both worlds.
Conceptually, {{SDF}} extracts image and point cloud feature, respectively, then performs sparse-fusion to aggregate geometric prior and dense-fusion to exploit semantic information.
The sparse-fusion module is responsible for gathering image feature around non-empty voxels, aiming to preserve the geometric prior provided by point cloud feature.
Meanwhile, the dense-fusion module projects point cloud feature to bird's-eye-view and queries image feature in perspective view, in the goal of utilizing rich semantic information in image.
Illustrated by Fig.~\ref{fig:motivation} (c), such a simple yet effective design can cover the disadvantage of sparse-only or dense-only fashion and enjoy the best of both worlds.
To demonstrate the effectiveness of our SDF design, we assemble two moderate methods, BEVFormer~\cite{li2022bevformer} for camera branch and TransFusion~\cite{bai2022transfusion} for LiDAR branch to construct the baseline network. Then the SDF framework improves their joint performance by a large margin, outperforms baseline by \textbf{4.3\%} in \textit{mAP} and \textbf{2.5\%} in NDS and reaches the state of the art on the nuScenes benchmark~\cite{caesar2020nuscenes}.

Extensive ablations respond to our analysis in the two fashions as well. Dense-only fashion has the advantage in texture-intensive categories such as truck and car, while sparse-only fashion is capable of localization-aware categories such as trailer and pedestrian.

To sum up, our work has three-fold contributions: 
\textbf{(a)} We propose a caveat analysis on existing sensor-fusion methods and point out their inevitable information loss respectively. 
\textbf{(b)} We devise a simple yet effective framework, named \textbf{{Sparse-Dense Fusion}}, which shares the advantage of sparse and dense fusion via a complementary structure.
\textbf{(c)} The proposed framework improves the baseline method by a large margin and ranks first on the nuScenes benchmark with extensive ablations.

\section{Related Work}

\subsection{3D Object Detection in General}

As autonomous driving system is a hierarchical design of perception~\cite{li2022delving}, prediction~\cite{jia2023towards} and planning~\cite{hu2023_uniad,wupolicy,hu2022st,wu2022trajectory}, 3D bounding boxes usually play the role of connecting the pipeline.
Based on the input modality, we divide 3D object detection research into three parts mainly - camera-based~\cite{wang2021fcos3d,wang2023object}, LiDAR-based~\cite{yin2021center} and sensor-fusion-based~\cite{liu2022bevfusion} methods.
Compared to the perspective view originating from camera, BEV representation has taken camera-based 3D perception by storm in terms of objection detection~\cite{li2022bevformer,huang2021bevdet}, lane detection~\cite{chen2022persformer,huang2023anchor3dlane} and BEV map segmentation~\cite{zhou2022cross}.
As BEV representation is a natural and straightforward candidate view to serve as a unified representation~\cite{li2022delving} across different modalities, representing image feature in BEV set up the foundation of sensor-fusion methods~\cite{liu2022bevfusion}.
In LiDAR-based methods, point cloud data is first processed to voxel~\cite{zhou2018voxelnet} or pillar~\cite{lang2019pointpillars} space, then go through 3D/2D feature extractor~\cite{yan2018second}, and finally, perform detection in BEV space~\cite{yin2021center,fu2021improved}.
Different from representing image feature in BEV, LiDAR-to-BEV transformation is more intuitive as point cloud data naturally preserve 3D geometric information. 

Given such progress in camera-based and LiDAR-based methods, fusing them to enjoy the best of both worlds is a natural idea, and many attempts did happen in the sensor-fusion domain.
In general, we present in detail various early- or mid-fusion methods in a taxonomy and categorize them into two fashions, termed \textbf{\textit{dense-only}} and \textbf{\textit{sparse-only}}.

\begin{figure}
  \centering
  \begin{overpic}[width=.45\textwidth]{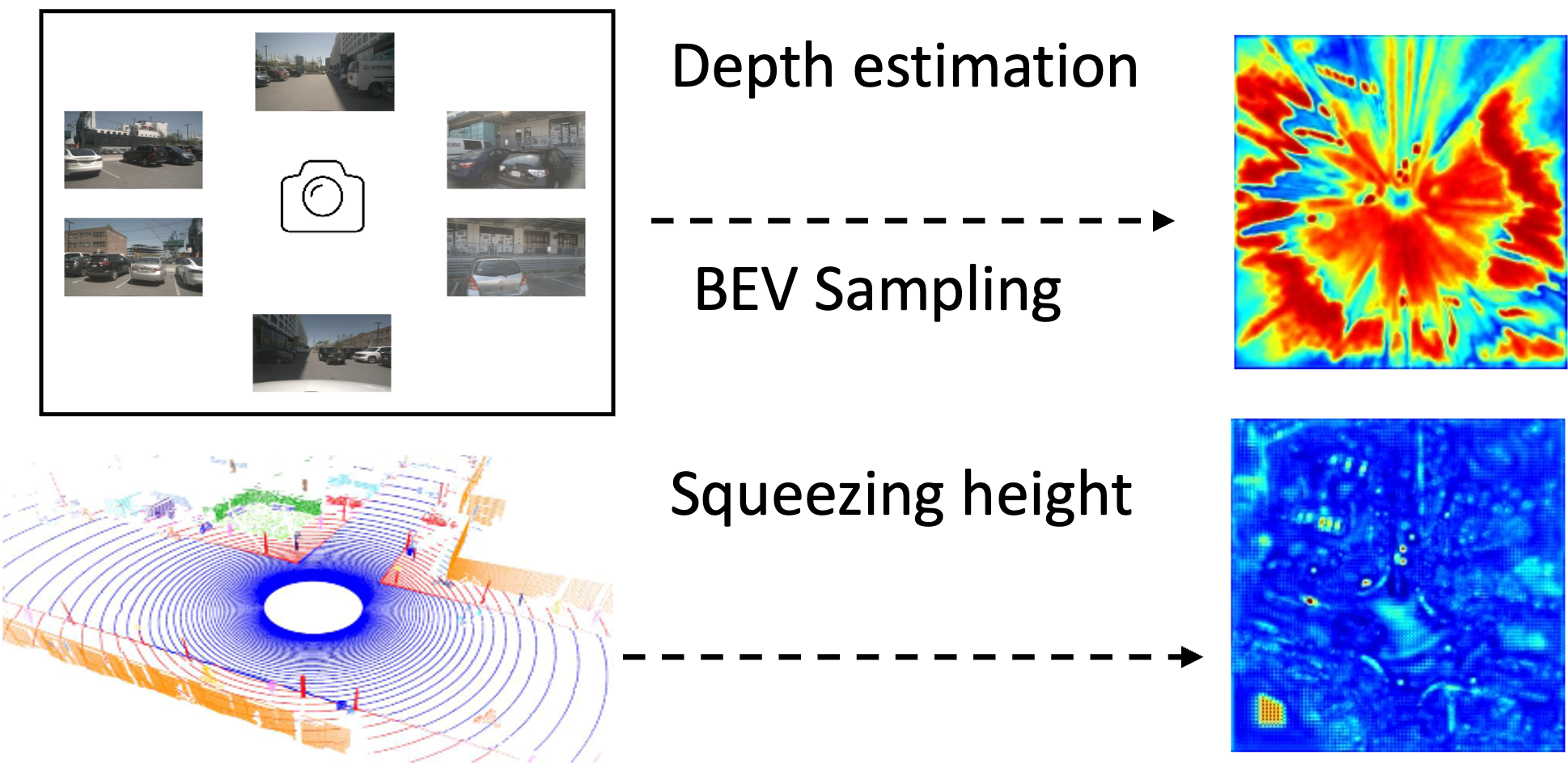}
    \put(43.5,24.9){\scriptsize~\cite{li2022bevformer,chen2022persformer,yang2022bevformer}}
    \put(41.5,37.9){\scriptsize~\cite{philion2020lift,li2022bevdepth,huang2021bevdet,liu2022bevfusion,liang2022bevfusion}}
    \put(43.5,11.4){\scriptsize~\cite{yin2021center,bai2022transfusion,li2022deepfusion,yan2018second}}
  \end{overpic}
  \caption{Visualization of \textbf{\textit{dense-only}} manner. Camera feature is transformed to BEV via depth estimation or BEV sampling, while LiDAR feature is via squeezing height in voxel space. It can be observed that there exists ``\textcolor{mypink1}{blurring long tail}'' in the shape of ray-tracing dilation. Since transforming camera feature to BEV is ill-posed, feature from two modalities cannot align well grid by grid, consequently hindering dense feature fusion.}
  \label{fig:dense}
  \vspace{-0.5cm}
\end{figure}

\subsection{Dense-only Fusion}

In dense-only fusion, image feature is either sampled to voxel space or projected to BEV space, and fused with point cloud feature in that representation.
Inspired by Lift-Splat-Shoot~\cite{philion2020lift}, projecting image feature to BEV space via depth estimation and simply concatenate with point cloud feature in BEV space have demonstrate its effectiveness by two BEVFusion~\cite{liu2022bevfusion, liang2022bevfusion}.
Similar approaches follow the idea of fusing feature in BEV space with additional tricks such as region of interest pooling in DeepInteraction~\cite{yang2022deepinteraction} and 3D-CVF~\cite{yoo20203d}, BEV space augmentation in DeepFusion~\cite{li2022deepfusion} and two-stage refinement in TransFusion~\cite{bai2022transfusion}.
Another line of works attempts to fuse feature in voxel space by expanding image feature to frustum and aligning it to voxel representation.
UVTR~\cite{li2022uvtr} transforms image feature to voxel space via predefined 3D point sampling, and aligns it with point cloud voxel feature.
Unlike BEV representation, voxel space preserve more 3D geometric information.

\subsection{Dense-only Fusion}

In sparse-only fusion, point cloud is projected to image plane and gather semantic label or image feature around the projection position.
In terms of gathering semantic labels, point-painting~\cite{vora2020pointpainting} and point-augmenting~\cite{wang2021pointaugmenting} project point cloud to image plane and append semantic labels around projection location to 3D points.
Recently the trend has been switched to fusing feature under this philosophy, such as AutoAlign~\cite{chen2022autoalign}, its V2~\cite{chen2022autoalignv2} and EPNet++~\cite{liu2021epnet++} focusing on the pixel-instance feature, MMF~\cite{liang2019multi} with multi-scale fusion, SFD~\cite{wu2022sparse} constructing pseudo point cloud from image, and VPFNet~\cite{zhu2022vpfnet} equipped with stereo data augmentation.
Another line of work applies object-level sparse fusion such as FUTR3D~\cite{chen2022futr3d} and CLOCS~\cite{pang2020clocs}.

\section{Analysis}

Though plenty of work has explored the potential of sensor-fusion following either sparse-only or dense-only fashion, both of them have inevitable information loss inherited from the design philosophy, respectively.
In this section, we will first make an ideal analysis of how a 3D object is perceived via different sensors.
Later, we highlight how the information loss happened in current sensor-fusion methods and finally introduce the motivation of our framework.
\subsection{Representation Matters in 3D Perception}
An object in the real world is captured by LiDAR into point cloud representation and by camera into image representation.
LiDAR is often regarded as a critical sensor for distance sensing in autonomous driving due to its ability to perceive and capture the 3D geometry of the environment.
The prevailing LiDAR configuration is equipped with a fixed number of beams, a default scanning frequency, and finally provides a stable number of scanned points per ring with slight fluctuation.
Consequently, the size of the point cloud data per frame is usually limited to tens of thousands, e.g., in nuScenes~\cite{caesar2020nuscenes}, approximately 35,000.
The camera is often seen as a complementary sensor for semantic information as it can capture the texture appearance of objects in the field of view ({FOV}).
Predominant camera configuration comprises several cameras with different locations and orientations on the autonomous vehicle. 
Each camera can perceive the scene in image with a resolution of, \textit{e.g.},  900$\times$1600 in nuScenes~\cite{caesar2020nuscenes}.
Therefore, in each frame, the camera system receives approximately 8,000,000 pixels in total.
The huge quantitative and qualitative difference between point cloud and image data implies the difficulty in designing sensor-fusion algorithms.

\begin{figure}
  \centering
    \includegraphics[width=.45\textwidth]{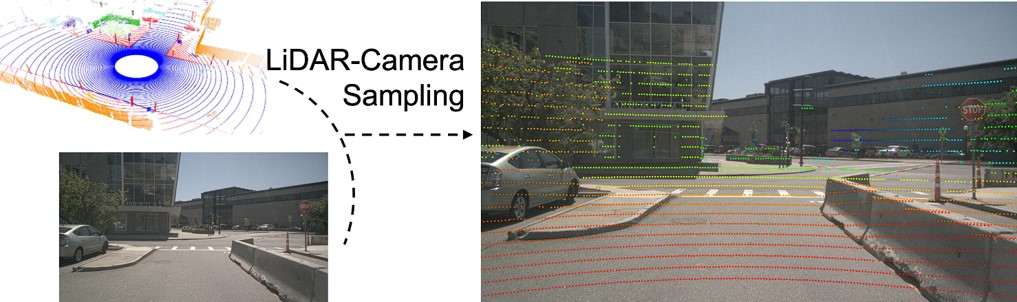}
  \caption{Visualization of \textbf{\textit{sparse-only}} manner. Point cloud is projected to image plane and gather semantic label or image feature around the projection position. Since the projection positions are sparse on image plane, the process of sampling feature will break the texture continuity in image.}
  \label{fig:sparse}
  \vspace{-0.25cm}
\end{figure}

\subsection{Achilles Heel of Sensor Fusion}
In a nutshell, the core problem of sensor fusion is to align the feature from different modalities, so transforming feature into the exact representation plays a vital role in feature.
Distinguished by where and how to transform feature, two intuitive methodologies have taken camera-LiDAR fusion by storm, the \textbf{\textit{sparse-only}} and the \textbf{\textit{dense-only}} fashion.

In sparse-only fusion, image feature is transformed to point cloud representation via LiDAR-camera sampling, and feature fusion operates only on those sampled positions.
The sampling process can be formulated as:
\begin{equation}\label{prel-3dv-E5}
        z_c \begin{bmatrix} u \\ v \\ 1 \end{bmatrix}
         = \begin{bmatrix}
            f_x & 0 & c_x \\
            0 & f_y & c_y \\
            0 & 0 & 1
            \end{bmatrix} 
        \begin{bmatrix}
        \textbf{R} & \textbf{T}
        \end{bmatrix}
        \begin{bmatrix}
        x_w & y_w & z_w & 1
        \end{bmatrix}^T , 
\end{equation}
where $(u,v)$ denotes 2D position in image coordinate, $z_c$ denotes the depth per pixel, $(x_w, y_w, z_w)$ denotes the 3D position in the world coordinate, $(f_x, f_y, c_x, c_y)$ denotes camera intrinsic, $(\textbf{R}~\textbf{T})$ denotes camera-LiDAR rotation and translation coefficient.
Such a transformation, however, breaks the texture continuity in image feature, and consequently cannot utilize the rich semantic information extracted via image backbone~\cite{he2016deep}.
Numerically, because of sparsity in point cloud, the sampled positions only count for around 1/100 to 1/1000 of the object appearance on the image. 
It's more evident through visualization in Fig.~\ref{fig:sparse}.
Therefore, sparse-only fashion has information loss in rich semantic feature provided by image.

In dense-only fusion, the image and point cloud features are transformed to BEV space, respectively, and are fused grid by grid.
Because BEV is a unified representation of different modalities, fusing features in BEV space is prevailing~\cite{liu2022bevfusion,liang2022bevfusion,yang2022deepinteraction}.

\begin{figure*}
  \centering
    \includegraphics[width=.95\textwidth]{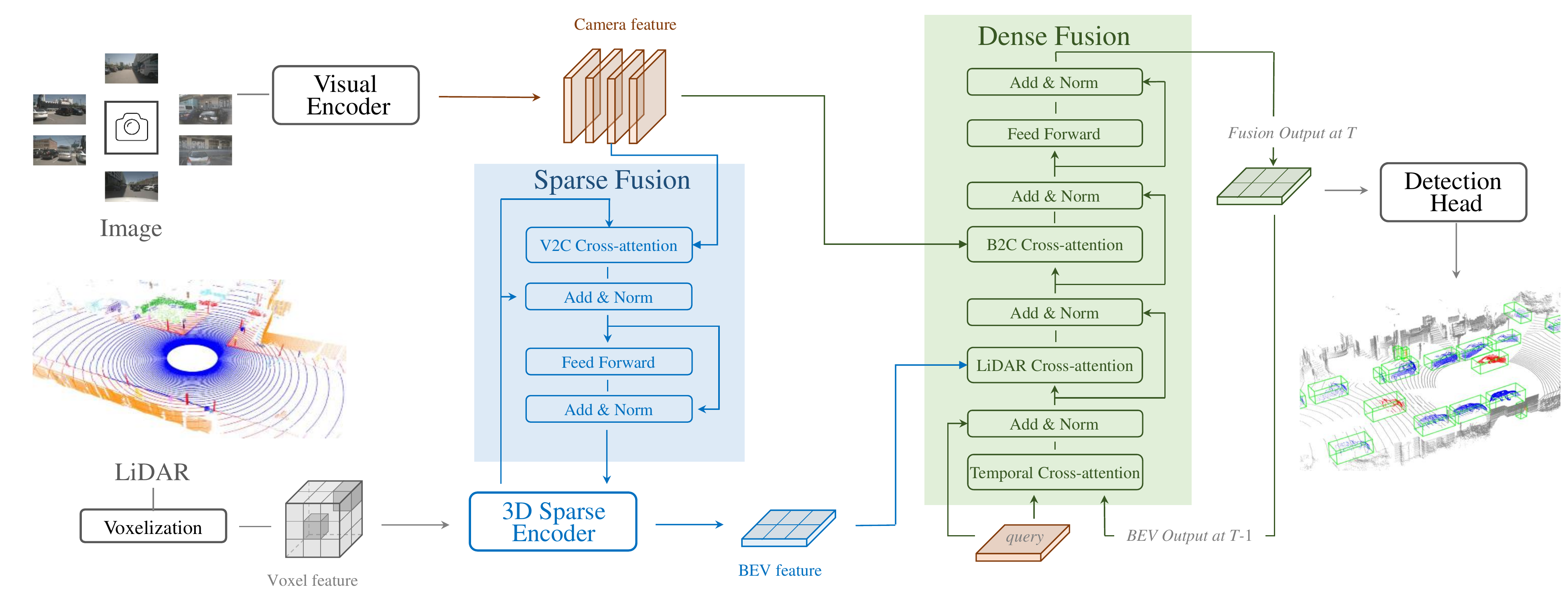}
  \caption{\textbf{Pipeline of our proposed SDF method.} SDF includes two sub-parts: Sparse Fusion Module and Dense Fusion Module. In Sparse Fusion Module, the non-empty voxel query interacts with image features. In Dense Fusion Module, each BEV query interacts alternately with the history BEV feature, lidar BEV feature, and image feature.
  }
  \label{fig:pipeline}
  \vspace{-0.5cm}
\end{figure*}

However, through compressing point cloud feature to BEV space, its 3D geometric structure is broken inevitably.
Meanwhile, projecting image feature to BEV space is ill-posed, resulting in location inconsistency.
Through visualization in Fig.~\ref{fig:dense}, it's observed that there exists ``blurring long tail" in camera-BEV feature because image does not preserve any 3D geometric prior, making the projection ill-posed.
Consequently, an object represented by several BEV grids in LiDAR-projected BEV feature may not align well with the corresponding BEV grids in camera-projected BEV feature.
Therefore, dense-only fashion has information loss in accurate 3D geometric feature provided by point cloud.

To this point, we have witnessed the strengths and weaknesses of each fashion; they are complementary to each other.
A straightforward idea is to incorporate both design schemes into a single network and enjoy the benefits from both worlds.

\section{Methodology}

This paper proposes a unified camera-lidar fusion framework based on attention. Our main contribution is fusing camera and lidar from both sparse and dense perspectives.

\subsection{Overall Structure}

The core idea of our SDF is fusing camera and lidar from both sparse and dense views. The overall architecture is illustrated in Fig.~\ref{fig:pipeline}. SDF includes two sub-parts: Sparse Fusion Module and Dense Fusion Module. 
Considering that deformable transformer~\cite{Zhu2021DeformableDD} has linear complexity and fast convergence speed, our module uses a deformable transformer~\cite{Zhu2021DeformableDD} to gather and fuse features in each part.
 Specifically, the Sparse Fusion Module fuses image features to valid voxels. Here, \textit{valid voxels} denote the voxels with lidar points inside. The fused voxel features maintain the original 3D spatial structure and are further sent to the subsequent network. As for dense fusion, we adopt BEV queries to integrate camera, lidar, and past BEV features, respectively. The BEV queries are grid-shaped parameters and are learned during training.

\begin{table*}[h]
    \centering
        \setlength{\aboverulesep}{0pt} 
        \setlength{\belowrulesep}{0pt} 
        \resizebox{\linewidth}{!}{
        \begin{tabular}{c|ccc|cc|cccccccccc}
        \toprule
          & \textbf{S-Fusion} & \textbf{D-Fusion} & \textbf{Temporal} & \textbf{mAP} & \textbf{NDS} & \scriptsize{Car} & \scriptsize{Truck} & \scriptsize{C.V.} & \scriptsize{Bus} & \scriptsize{Trailer} & \scriptsize{Barrier} & \scriptsize{Motor.} & \scriptsize{Bike} & \scriptsize{Ped.} & \scriptsize{T.C.}\\
        \hline
        1  &         &     &         & 64.9   & 69.9  &86.9 &60.9 &26.1 &73.7 &42.8 &68.8 &71.3 &55.9 &86.8 &74.4   \\
        2  & \cmark &     &         & 68.1(+3.2)   & 70.9(+1.0)  &88.4  &64.4 &30.0 &77.2 &45.8 &69.4 &77.0 &63.1 &88.1 &77.5            \\
        3  &       & \cmark &       & 67.9(+3.0)   & 71.5(+1.6)  &89.4 &65.8 &28.6  &77.4 &43.8 &70.6 &76.9 &61.2 &87.7 &77.7             \\
        \rowcolor[RGB]{220,220,220} 4  & \cmark &  \cmark &    & 68.8(+3.9)   & 72.0(+2.1) &89.5 &65.9 &30.7 &77.6	&45.2 &70.6 &78.2 &63.5 &88.3 &77.9             \\
        \rowcolor[RGB]{220,220,220} 5  & \cmark &  \cmark & \cmark & \textbf{69.2}(+4.3)  & \textbf{72.4}(+2.5)  &89.5 &67.0 &32.4 &77.3	&43.4 &71.5 &79.2 &65.7 &88.5 &77.8    \\
        \bottomrule
        
        \end{tabular}
    }
    \caption{\textbf{Component analysis.} We verify the effectiveness of the Sparse Fusion Module and Dense Fusion Module. ``S-Fusion'' and ``D-Fusion'' denotes the Sparse and Dense Fusion Module, respectively. ``Temporal" means the Dense Fusion Module with temporal information. ``C.V.", ``Ped.", and ``T.C." are abbreviations of construction vehicle, pedestrian, and traffic cone, respectively. Compared with each other, sparse-only fashion performs better on small objects such as bikes and pedestrians, while dense-only fashion performs better on objects with large scale and regular locations such as cars, trunks, and buses. Sparse Dense Fusion (SDF) benefits from both.}
    \label{tb:components}
    \vspace{-0.25cm}
\end{table*}

\subsection{Sparse Fusion in Voxel Space}
\label{sec:sparse_fusion}

The purpose of sparse fusion is to preserve 3D geometric prior in the fusion process. In the pipeline of the entire lidar branch, voxel space can maintain geometric information. Since the voxel space is large and sparse, taking all voxel as queries for fusion is useless and inefficient. Thus, we propose a modified deformable attention module to establish valid connections between the voxel and camera features.

Specifically, we focus on valid voxels to improve efficiency and accuracy. We take the centers of all valid voxels as 3D reference points and project them onto image planes to get 2D reference points. 
Then, we use V2C(voxel-to-camera) cross-attention to gather the camera feature, defined as:

\begin{equation}
\label{sca}    
    \text{V2C}(Q_p, F) = \frac{1}{|\mathcal{V}_\text{hit}|} \sum_{i\in \mathcal{V}_\text{hit}} \sum_{j=1}^{{N_\text{ref}}}
    \text{DeAttn}(Q_p, \mathcal{P}(p,i,j), F^i),
\end{equation}

\noindent where $\mathcal{V}_\text{hit}$ indexes the number of images hit by 2D reference points, $i$ indexes the image view, $j$ indexes the reference points, and ${N_\text{ref}}$ is the total number of reference points for each voxel query. $F^i$ is the features of the $i$-th camera view. For each voxel query $Q_p$, we utilize a project function $\mathcal{P}(p, i, j)$ to get the $j$-th reference point on the $i$-th view image.

Following that, we show how to build 2D reference points on a view image using the projection function $\mathcal{P}$ and how to adapt to lidar geometric augmentation. We calculate the real world position $(x', y', z')$ corresponding to the query $Q_p$ located at $p=(x,y,z)$ of $Q$ as follows:
\begin{equation}{\label{eqn:real}}
        x'\!=\! (x\!-\!\frac{W}{2})\!\times\! s_x; \quad y'\! = \!(y\!-\!\frac{H}{2})\!\times\! s_y;  \quad z'\! = \!(z\!-{Z})\!\times\! s_z + {z_{base}},
\end{equation}
where $H$, $W$, $Z$ are the spatial shapes of the entire  voxel feature, $s_x$, $s_y$, $s_z$ are the size of resolution of voxel, ${z_{base}}$ is the base height, and $(x',y', z')$ are the coordinate of the ego vehicle coordinate system.
As a result, for each query $Q_p$, we obtain 3D reference points ${(x', y', z')}$. Finally, we project the 3D reference points to different image views using Eqn.~\ref{prel-3dv-E5}.
When we apply data augmentation to a point cloud, we reverse all 3D reference points according to data augmentation, and then use the projection matrix to project the 3D reference points to 2D reference points.

\subsection{Dense Fusion in BEV Space}

Our Dense Fusion Module is composed of 6 transformer layers, and each transformer layer is composed of Temporal Cross-attention, LiDAR Cross-attention, B2C(Bev to Camera) Cross-attention module, and Feed Forward module. The purpose of these cross-attention modules is to gather temporary information, lidar information, and image information, respectively. In the beginning, we initialize the BEV query following BEVFormer~\cite{li2022bevformer}.

Different from the sparse fusion, which only performs fusion on valid voxel, Dense Fusion Module treats each BEV query equally. First, we initialize the BEV query with a set of learnable parameters. Then, the BEV queries are used to collect and integrate features from different modalities. Directly fusion in BEV space will cause semantic ambiguity because of the height compression. Following BEVFormer~\cite{li2022bevformer}, we predefine a set of anchor heights for each BEV query to alleviate this problem.

When fusing camera features, we set 3D reference points for the BEV query at different heights and use the same projection method described in section~\ref{sec:sparse_fusion} to generate the 2D reference points. As for lidar fusion, we directly set reference points to gather lidar features at the corresponding locations. 

As shown in Fig.~\ref{fig:pipeline}, our Dense Fusion Module contains a modified transformer layer that alternately fuses the lidar, camera, and history BEV features. The BEV features generated by the Dense Fusion Module will be sent to the 3D detection head. In contrast to Temporal Cross-attention, we first align $B_{t\!-\!1}$ to $Q$ according to ego-motion, then update the 2D reference point location to keep the features at the same grid corresponding to the exact real-world location. With the above modification, we get the final BEV feature equipped with temporal and multimodal information.
\section{Experiments}

In this section, we conduct extensive comparisons and ablation studies to examine the effectiveness of our method. 
We use the gray background to indicate our method in the following tables.

\subsection{Setup and Implementation Details}

\textbf{Dataset.} We conduct experiments on the nuScenes dataset~\cite{caesar2020nuscenes}, which contains 1000 scenes, with each sample consisting of RGB images from 6 cameras with 360° horizontal FOV and one LiDAR point cloud. There are 1.4 million 3D bounding boxes with annotations from ten categories.
We use mean average precision (mAP) and nuScenes detection score (NDS) evaluation metrics provided by the dataset to evaluate our results.

\textbf{Implementation details.} 
We implement SDF with Python 3.8.13, PyTorch 1.9.0, and CUDA 11.3. All models are trained with 8 Nvidia A100 GPUs. 
Following BEVFormer~\cite{li2022bevformer}, we use ResNet101-DCN~\cite{he2016deep} as our image backbone. For lidar, we use the VoxelNet~\cite{zhou2018voxelnet} as our LiDAR backbone, and TransFusion-L~\cite{bai2022transfusion} is chosen as our 3D detection head. And we also conduct a two-stage training process following BEVFusion~\cite{liang2022bevfusion}.
Like BEVFusion, we do not use data augmentation when training the camera and fusion models. And we only use data augmentation for the fusion model when comparing it with the state-of-the-art methods. And we employ a single model during testing with no test-time augmentation.

\begin{figure*}[t]
\centering
\includegraphics[width=\linewidth]{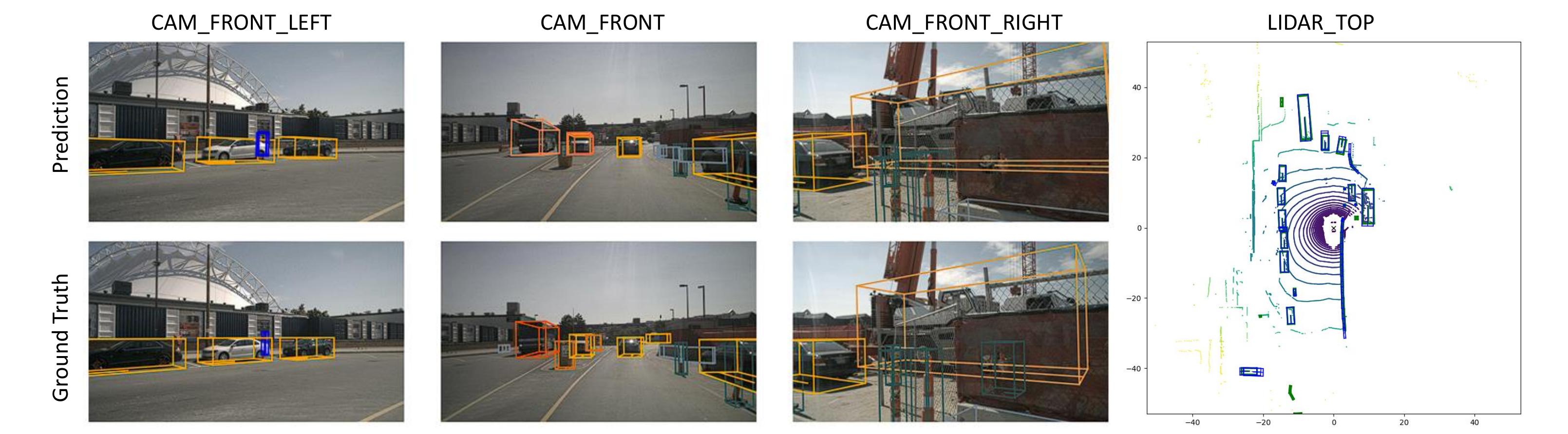}
\caption{\textbf{Visualization results of SDF on nuScenes \texttt{val} set.} We show the 3D bounding box predictions across each camera's field of view and the bird-eye view. Except for a few errors in crowded areas and remote objects, our method yields remarkable results.
}
\label{fig:vis}
\vspace{-0.25cm}
\end{figure*}

\subsection{Components of SDF}
We examine the effectiveness of each proposed module, and results are listed in Tab.~\ref{tb:components}.
The \nth{1} row is our baseline Transfusion-L~\cite{bai2022transfusion}. 
We individually add Sparse Fusion Module and Dense Fusion Module to the baseline model, and the result is shown in the \nth{2} and \nth{3} rows.
Each component improves the performance, indicating their effectiveness in the 3D detection task. 
In particular, we can observe that sparse fusion has better performance on small objects such as bikes and pedestrians, which proves that sparse fusion can effectively leverage the spatial structure and has a more accurate mapping from 3D space to camera view than dense fusion.
And dense fusion performs better on objects with large scale and regular location such as car, trunk, and bus, because such objects rely less on spatial structure and dense fusion provides richer semantic information.
The \nth{4} row is the result of combining both fusion modules, achieving 4.0\% mAP and 2.1\% NDS absolute improvements compared to the baseline. Besides, using both fusion modules is better than using only one, showing that our sparse and dense modules can complement each other. Additionally, in the \nth{5} row, we extend our model in the temporal dimension, and the result shows that our model can effectively utilize the temporal information.

\begin{table}[ht!]
    \centering
        \setlength{\aboverulesep}{0pt} 
        \setlength{\belowrulesep}{0pt} 
        \resizebox{\linewidth}{!}{
        \begin{tabular}{c|c|c|cc}
        \toprule
          & \textbf{Loc} & \textbf{Method} & \textbf{mAP} & \textbf{NDS} \\
        \hline
        1      & $C_1$      & deformable attention                        & 66.6            & 70.4     \\
        2      & $C_2$      & deformable attention                       &66.9             & 70.0              \\
        3       & $C_3$     & deformable attention                        & 67.3    & 70.8              \\
        \rowcolor[RGB]{220,220,220}4       & $C_4$     &  deformable attention                        & \textbf{68.1}    & 70.9              \\
        5       & $C_5$     &  deformable attention                        & 67.4    & 70.4              \\
        6       & $C_4$     &  extended deformable attention               & \textbf{68.1}          & \textbf{71.2}\\

        \bottomrule
        
        \end{tabular}
    }
    \caption{Sparse Fusion Module analysis. We explore the influence of different fusion locations and different fusion methods in the sparse module. “Loc” denotes the location of the Sparse Fusion Module. $C_x$ means insert the deformable attention after layer x of the sparse encode. 
    }
    \label{tb:sparse_module}
    \vspace{-0.25cm}
\end{table}

\begin{table*}[t]
\centering
\setlength{\tabcolsep}{4.0pt}
\resizebox{0.95\linewidth}{!}{
\begin{tabular}{l|cc|cc|cccccccccc}
    \toprule
         \hspace{-0.3em}{Method}  & Modality & Fusion Type & mAP & NDS & \scriptsize{Car} & \scriptsize{Truck} & \scriptsize{C.V.} & \scriptsize{Bus} & \scriptsize{Trailer} & \scriptsize{Barrier} & \scriptsize{Motor.} & \scriptsize{Bike} & \scriptsize{Ped.} & \scriptsize{T.C.} \\
         \midrule
        \hspace{-0.3em}\texttt{\textcolor{gray}{val}} & & & & & & & & & & & & & \\
        {FUTR3D}~\cite{chen2022futr3d} &LC &D &64.2 &68.0  &86.3  &61.5  &26.0  &71.9  &42.1  & 64.4 & 73.6 & 63.3 &82.6 &70.1 \\
         {BEVFusion*}~\cite{liang2022bevfusion} &LC &D &69.6 &72.1 &89.1 &66.7 &30.9 &77.7 &42.6 &73.5 &79.0 &67.5 &89.4 &79.3\\
         {BEVFusion}~\cite{liu2022bevfusion} &LC &D &68.5 &71.4 &- &- &- &-	&- &- &- &- &- &-\\
         {Deepinteraction}~\cite{yang2022deepinteraction} &LC &D &69.9 &72.6 &- &- &- &-	&- &- &- &- &- &-\\
         \rowcolor[RGB]{220,220,220}{Ours} &LC &SD &68.8 &72.0 &89.5 &65.9 &30.7 &77.6	&45.2 &70.6 &78.2 &63.5 &88.3 &77.9\\

         \rowcolor[RGB]{220,220,220}{Ours-T} &LC &SD &69.2 &72.4 &89.5 &67.0 &32.4 &77.3	&43.4 &71.5 &79.2 &65.7 &88.5 &77.8\\

         \rowcolor[RGB]{220,220,220}{Ours*} &LC &SD &70.4 &72.8 &90.0 &68.2 &33.5 &79.5	&48.3 &70.8 &78.4 &66.5 &89.1 &79.4\\
        \midrule
        \hspace{-0.3em}\texttt{\textcolor{gray}{test}} & & & & & & & & & & & & & \\
        {PointPillars~\cite{lang2019pointpillars}} & L &-  & 30.5 & 45.3 & 68.4 & 23.0 & 4.1 & 28.2 & 23.4 & 38.9 & 27.4 & 1.1 & 59.7 & 30.8 \\
          {CenterPoint$^\dag$~\cite{Yin2020Centerbased3O}} & L &-  & 60.3 & 67.3 & 85.2 & 53.5 & 20.0 & 63.6 & 56.0 & 71.1 & 59.5 & 30.7 & 84.6 & 78.4 \\
         {TransFusion-L~\cite{bai2022transfusion}} & L &-  & {65.5} & {70.2} & {86.2} & {56.7} & {28.2} & {66.3} & {58.8} & {78.2} & {68.3} & {44.2} & {86.1} & {82.0} \\
          {PointPainting~\cite{vora2020pointpainting}} & LC &S   & 46.4 & 58.1 & 77.9 & 35.8 & 15.8 & 36.2 & 37.3 & 60.2 & 41.5 & 24.1 & 73.3 & 62.4 \\
          {3D-CVF~\cite{yoo20203d}} & LC &D   & 52.7 & 62.3 & 83.0 & 45.0 & 15.9 & 48.8 & 49.6 & 65.9 & 51.2 & 30.4 & 74.2 & 62.9 \\
         {TransFusion~\cite{bai2022transfusion}} & LC &D  & {68.9} & {71.7} & 87.1 & 60.0 & {33.1} & 68.3 & 60.8 & {78.1} & 73.6 & {52.9} & {88.4} & {86.7} \\
         {Autoalignv2}~\cite{chen2022autoalignv2}  &LC &S &{68.4} &{72.4} 
         &{87.0} &{59.0} &{33.1} &{69.3} &{59.3}  &{78.0}  &72.9 &52.1 &{87.6} &{85.1} \\
         {BEVFusion*}~\cite{liang2022bevfusion}  &LC &D &{71.3} &{73.3} 
         &{88.5} &{63.1} &{38.1} &{72.0} &{64.7}  &{78.3}  &75.2 &56.5 &{90.0} &{86.5} \\
         {BEVFusion}~\cite{liu2022bevfusion}  &LC &D &{70.2} &{72.9} 
         &{88.6} &{60.1} &{39.3} &{69.8} &{63.8}  &{80.0}  &74.1 &51.0 &{89.2} &{86.5} \\
        {DeepInteraction}~\cite{yang2022deepinteraction}  &LC &D &{70.8} &{73.4} 
         &{87.9} &{60.2} &{37.5} &{70.8} &{63.8}  &{80.4}  &75.4 &54.5 &{91.7} &{87.2} \\

         \rowcolor[RGB]{220,220,220} {Ours*}  &LC &SD &{70.9} &{73.4} 
         &{89.2} &{63.2} &{38.6} &{73.2} &{65.5}  &{80.9}  &75.3 &48.1 &{89.2} &{86.0} \\

        \bottomrule
	\end{tabular}
 }
	\caption{\textbf{Performance on the nuScenes 
 \texttt{val} and \texttt{test} sets}. ``C.V.", ``Ped.", and ``T.C." are abbreviations of construction vehicle, pedestrian, and traffic cone, respectively. `L' and `C' represent LiDAR and camera, respectively. `D', `S', and `SD' represent `Dense-only', `Sparse-only', and 'Sparse Dense'. $\dag$ indicates the results with double-flip during the test. * indicates method with data augmentation during training. Even \textit{without} temporal information, our SDF achieves new state-of-the-art on both \texttt{val} and \texttt{test} sets. Particularly, the augmentation version of SDF achieves 0.8\% and 0.6\% absolute improvements compared to BEVFusion with augmentation during training.
 }
	\label{tab:compare_to_sota}
	\vspace{-0.1cm}
\end{table*}

\subsection{Ablation Study}

\textbf{Robustness.} 
Here, we simulate LiDAR sensor failures to test the robustness of different methods. Following BEVFusion~\cite{liang2022bevfusion}, we set a limited field-of-view (FOV) and remove the out-of-view LiDAR points. 
Tab.~\ref{tb:robustness} shows the influence of limited LiDAR field-of-view. One can observe that the Sparse-only method is slightly better than the LiDAR-only method. The Sparse-only method relies heavily on the LiDAR point, so the fusion process can only bring slight improvement compared to the LiDAR-only method. The Dense-only method is much better than the LiDAR-only method, which achieves 21.3\% and 12.7\% absolute improvements on mAP and NDS, respectively, under the robustness setting.

\begin{table}[h]
    \centering
        \setlength{\aboverulesep}{0pt} 
        \setlength{\belowrulesep}{0pt}  
        \begin{tabular}{c|c|c|cc}
        \toprule
           & \textbf{Fusion type} & \textbf{FOV} & \textbf{mAP} & \textbf{NDS} \\
        \hline
        1            & - & -                         &  64.9           & 69.9    \\
        2            & - & ($-\pi/2$, $\pi/2$)                        &  28.1           & 50.6     \\

        3            & S & -                     &  68.1            &   70.9    \\
        4            & S &  ($-\pi/2$, $\pi/2$)                   & 30.5             &  51.4             \\
        5            & D  &  -                  &  67.9   & 71.5              \\
        6            & D  &  ($-\pi/2$, $\pi/2$)                   &  49.4   &   58.2            \\
        \rowcolor[RGB]{220,220,220} 7            & SD  & -                   & 68.8     &    72.0           \\
        \rowcolor[RGB]{220,220,220} 8            & SD  & ($-\pi/2$, $\pi/2$)                    & 50.2    & 59.0               \\

        \bottomrule
        
        \end{tabular}
    \caption{The result of limited LiDAR field-of-view. Dense-only Fusion outperforms baseline by 21.3\%mAP and 12.7\% NDS under the robustness setting, which proves its robustness.}
    \label{tb:robustness}
    \vspace{-0.25cm}
\end{table}

\textbf{Efficiency.} 
We compare the efficiency of the Sparse Fusion Module and Dense Fusion Module in the fusion process by calculating the number of fusion blocks~(voxels or BEV grids). 
Because the number of valid voxels is related to the input point cloud, here we count the average of valid voxels after the fourth layer of the sparse encoder in Val set . 
The dense Fusion Module fuses on all grids, so the number of fusion grids depends on the size of the BEV. In our experiment, the BEV size is set to 180$\times$180, so the number of fusion grids is 32400.
As shown in Tab.~\ref{tb:efficiency}, the number of fusion blocks for our Sparse Fusion Method is approximately 13k, fewer than the Dense Fusion Method, which proves that sparse fusion is more efficient than dense fusion. 

\begin{table}[h]
    \centering
        \setlength{\aboverulesep}{0pt} 
        \setlength{\belowrulesep}{0pt}  
        \begin{tabular}{c|c|c|cc}
        \toprule
           & \textbf{Fusion type} & \textbf{NF} & \textbf{mAP} & \textbf{NDS} \\
        \hline
        1            & - &  -                    & 64.9   & 69.9      \\
        2            & S &  13k                  & 68.1  & 70.9              \\
        3            & D  & 32k             & 67.9   & 71.5              \\
        \rowcolor[RGB]{220,220,220} 4            & SD  & 55k                & \textbf{68.8}   & \textbf{72.0}              \\

        \bottomrule
        
        \end{tabular}
    \caption{The number of fusion blocks. By calculating the number of fusion blocks, we can find that Sparse-only Fusion has fewer fusion blocks than Dense-only Fusion.}
    \label{tb:efficiency}
    \vspace{-0.25cm}
\end{table}

\textbf{Sparse Fusion Module.} 
We then explore different settings of the sparse module and summarize the results in Tab.~\ref{tb:sparse_module}. 
From the \nth{1} to \nth{5} row, we insert our V2C deformable attention into the different locations of the sparse encoder. It is effective to insert the deformable attention after the fourth layer of the sparse encoder. In other words, deeper voxel features are more beneficial for aggregating image information. 
Furthermore, we conduct experiments on the use of reference points. In the \nth{6} row, we use offset sampling while setting multiple reference points to voxels adjacent to the query.
The modification (Row 6) performs slightly better than the original deformable attention (Row 4) because of its wider perception field. Nevertheless, we still use the original V2C deformable attention to keep it simple.

\begin{table}[h]
    \centering
        \setlength{\aboverulesep}{0pt} 
        \setlength{\belowrulesep}{0pt} 
        \resizebox{\linewidth}{!}{
        \begin{tabular}{c|c|c|cc}
        \toprule
           & \textbf{Method} & \textbf{\# Encoder Layer} & \textbf{mAP} & \textbf{NDS} \\
        \hline
        1            & dynamic fusion & 6                     & 66.9             & 70.9              \\
        2            & fusion encoder  & 1                    & 66.8    & 70.8              \\
        \rowcolor[RGB]{220,220,220}3            & fusion encoder  & 6                     & \textbf{67.9}    & \textbf{71.5}              \\
        \bottomrule
        
        \end{tabular}
    }
    \caption{Dense Fusion Module analysis. We explore the impact of different fusion methods and different numbers of encoder layers. 
    }
    \label{tb:dense_module}
\end{table}

\textbf{Dense Fusion Module.} 
We compare our fusion method in the dense module with dynamic fusion from BEVFusion~\cite{liang2022bevfusion}. Dynamic fusion first concatenates multimodal features and fuses them with learnable static weights, then applies channel attention to select fused features. This module is simple but lacks the ability for iterative interaction between features. To apply the dynamic fusion module in our framework, we remove the lidar input in the dense fusion encoder to get the pure BEV camera feature and then fuse the lidar and camera BEV features with the dynamic fusion module.
As shown in Tab.~\ref{tb:dense_module}, our dense fusion method is better than dynamic fusion when using the same number of encoders. With iterative interaction, our fusion encoder achieves more sufficient fusion. 
We also test the scalability of our Dense Fusion Module and conclude that more encoder layers can lead to better results.

\begin{table}[h]
    \centering
        \setlength{\aboverulesep}{0pt} 
        \setlength{\belowrulesep}{0pt}  
        \begin{tabular}{c|c|c|cc}
        \toprule
           & \textbf{Method} & \textbf{SD-Fusion} & \textbf{mAP} & \textbf{NDS} \\
        \hline
        1            & Centerpoint &                         & 57.1            & 65.4     \\
        2            & Centerpoint & \cmark                    & 63.9             & 69.2              \\
        3            & Transfusion-L  &                    & 64.9    & 69.9              \\
        \rowcolor[RGB]{220,220,220} 4            & Transfusion-L  & \cmark                    & \textbf{68.8}    & \textbf{72.0}              \\
        \bottomrule
        
        \end{tabular}
    \caption{Expandability  analysis. With our Sparse Dense Fusion,  Centerpoint and Transfusion-L get 6.8\% 3.8\% and 3.9\%, 2.8\% absolute improvements on mAP and NDS respectively.}
    \label{tb:head}
    \vspace{-0.25cm}
\end{table}

\textbf{Different Detection Methods.}
Tab.~\ref{tb:head} shows the result of different detection methods with our fusion module. 
Our fusion method performs feature fusion at the voxel or BEV space, which can be easily extended to existing detection methods. With the SD-Fusion, Centerpoint and Transfusion-L get 6.8\% 3.8\% and 3.9\%, 2.8\% absolute improvements on mAP and NDS respectively, indicating that our approach has the general ability to apply to different detection methods.

\begin{table}[h]
    \centering
        \setlength{\aboverulesep}{0pt} 
        \setlength{\belowrulesep}{0pt}  
        \begin{tabular}{c|c|ccc}
        \toprule
           & \textbf{\#Frame}  & \textbf{mAP} & \textbf{NDS} &\textbf{mAVE} \\
        \hline
        1             &   1                      & 68.8            & 72.0  &  26.2  \\
        2             & 2                    & 68.7             & 72.0    &   25.8       \\
        \rowcolor[RGB]{220,220,220} 3             &  3                   & \textbf{69.2}    & \textbf{72.4}   &    \textbf{25.5}       \\
        \bottomrule
        
        \end{tabular}
    \caption{Temporal analysis. We explore the influence of different frame numbers during training. “\#Frame” denotes the frame number during training.
    }
    \label{tb:temporal}
    \vspace{-0.5cm}
\end{table}

\textbf{Different frame number during training.}
Tab.~\ref{tb:temporal} shows the effect of the frame number during training. 
We can observe that introducing temporal information could bring improvements to our model. Particularly, our method with 3 frames during training achieves 0.4\% and 0.4\% absolute improvements compared to the version that does not leverage temporal information.

\subsection{Comparison to State-of-the-arts} \label{sec: sota}

In Tab.~\ref{tab:compare_to_sota}, we report our main results on nuScenes \texttt{val} and \texttt{test} splits. On the \texttt{val} set, our method achieves the state-of-the-art performance of 70.4\%mAP and 72.7\%NDS, outperforming all previous methods.
Compared to BEVFusion, which is a dense-only fashion, our method can surpass it in terms of both non-augmentation and augmentation versions. In particular, the augmentation version of our method achieves 0.8\% and 0.6\% absolute improvements compared to BEVFusion.
On the \texttt{test} set, our method achieves state-of-the-art with 70.9\% mAP and 73.4\% NDS. 

\textbf{Visualization}
We show the detection results of a complex scene in Fig.~\ref{fig:vis}. 
Except for a few errors in crowded areas and remote objects, our method yields remarkable results.

\section{Conclusion}

In this paper, we analyze current fusion methods and point out their inevitable information loss.
Thus, we propose a unified framework termed Sparse-Dense Fusion, which enriches semantic texture and exploits spatial structure. 
The framework significantly outperforms the baseline method and ranks first on the nuScenes benchmark.








\bibliography{egbib.bib}

\bibliographystyle{IEEEtran.bst}

\end{document}